\documentclass{article} 
\usepackage{iclr2019_conference,times}

\usepackage{hyperref}
\usepackage{url}

\usepackage{graphicx}

\usepackage{amsmath,amssymb}
\usepackage{color}

\usepackage{algorithmic}
\usepackage{algorithm}

\usepackage{multirow}
\usepackage{makecell} 

\DeclareMathOperator{\sign}{sign}
\DeclareMathOperator{\sech}{sech}
\newcommand{\thscale}{\nu}

\newcommand{\W}{\mathbf{W}}
\renewcommand{\P}{\mathbf{P}}
\newcommand{\B}{\mathbf{B}}

\newcommand{\A}{\mathbf{A}}
\renewcommand{\O}{\mathbf{O}}
\newcommand{\BC}{\textbf{BC}}
\newcommand{\BNN}{\textbf{BNN}}
\newcommand{\BWN}{\textbf{BWN}}
\newcommand{\XNOR}{\textbf{XNOR}}
\newcommand{\BN}{\emph{BN}}
\newcommand{\SBN}{\emph{SBN}}
\newcommand{\BinaryBN}{\emph{BinaryBN}}

\title{Self-Binarizing Networks}

\author{Fayez Lahoud$^1$, Radhakrishna Achanta$^2$, Pablo M\'arquez-Neila$^3$, and Sabine S{\"u}sstrunk$^1$  \\
$^1$School of Computer and Communication Sciences, EPFL\\
$^2$Swiss Data Science Center, EPFL\\
$^3$ARTORG Center for Biomedical Engineering Research, University of Berne\\
\texttt{fayez.lahoud@epfl.ch}\\
\texttt{radhakrishna.achanta@datascience.ch}\\
\texttt{pablo.marquez@artorg.unibe.ch}\\
\texttt{sabine.susstrunk@epfl.ch}\\
}

\newif\ifdraft
\draftfalse

\definecolor{orange}{rgb}{1,0.5,0}
\definecolor{pink}{rgb}{0.98, 0.38, 0.5}

\ifdraft
 \newcommand{\PMN}[1]{{\color{orange}{\bf PMN: #1}}}
 
\else
 \newcommand{\PMN}[1]{{\color{red}{}}}
 
\fi

\ifdraft
 \newcommand{\RK}[1]{{\color{purple}{\bf PMN: #1}}}
 
\else
 \newcommand{\RK}[1]{{\color{red}{}}}
 
\fi

\iclrfinalcopy 
\begin{document}

\maketitle

\begin{abstract}

We present a method to train self-binarizing neural networks, that is, networks that evolve their weights and activations during training to become binary. To obtain similar binary networks, existing methods rely on the sign activation function. This function, however, has no gradients for non-zero values, which makes standard backpropagation impossible. To circumvent the difficulty of training a network relying on the sign activation function, these methods alternate between floating-point and binary representations of the network during training, which is sub-optimal and inefficient. We approach the binarization task by training on a unique representation involving a smooth activation function, which is iteratively sharpened during training until it becomes a binary representation equivalent to the sign activation function. Additionally, we introduce a new technique to perform binary batch normalization that simplifies the conventional batch normalization by transforming it into a simple comparison operation. This is unlike existing methods, which are forced to the retain the conventional floating-point-based batch normalization. Our binary networks, apart from displaying advantages of lower memory and computation as compared to conventional floating-point and binary networks, also show higher classification accuracy than existing state-of-the-art methods on multiple benchmark datasets.

\end{abstract}

\section{Introduction}
\label{sec:one}

Deep learning has brought about remarkable advancements to the state-of-the-art in several fields including computer vision and natural language processing. In particular, convolutional neural networks (CNN's) have shown state-of-the-art performance in several tasks such as object recognition with AlexNet~\citep{krizhevsky2012imagenet}, VGG~\citep{simonyan2014very}, ResNet~\citep{he2016deep} and detection with R-CNN~\citep{girshick2014rich,girshick2015fast,ren2017faster}. However, to achieve real-time performance these networks are dependent on specialized hardware like GPU's because they are computation and memory demanding. For example, AlexNet takes up 250Mb for its 60M parameters while VGG requires 528Mb for its 138M parameters.

While the performance of deep networks has been gradually improving over the last few years, their computational speed has been steadily decreasing~\citep{vedaldi16deepcv}. Notwithstanding this, interest has grown significantly in the deployment of CNN's in virtual reality headsets (Oculus, GearVR), augmented reality gear (HoloLens, Epson Moverio), and other wearable, mobile, and embedded devices. While such devices typically have very restricted power and memory capacites, they demand low latency and real-time performance to be able to provide a good user experience. Not surprisingly, there is considerable interest in making deep learning models computationally efficient to better suit such devices~\citep{ota_survey_2017,cheng_survey_2018,wang_survey_2018}.

Several methods of compression, quantization, and dimensionality reduction have been introduced to lower memory and computation requirements. These methods produce near state-of-the-art results, either with fewer parameters or with lower precision parameters, which is possible thanks to the redundancies in deep networks~\citep{cheng2015exploration}.

In this paper we focus on the solution involving binarization of weights and activations, which is the most extreme form of quantization. Binarized neural networks achieve high memory and computational efficiency while keeping performance comparable to their floating point counterparts. ~\cite{courbariaux2015binaryconnect} have shown that binary networks allow the replacement of multiplication and additions by simple bit-wise operations, which are both time and power efficient.

The challenge in training a binary neural network is to convert all its parameters from the continuous domain to a binary representation, typically done using the $\sign$ activation function. However, since the gradient of $\sign$ is zero for all nonzero inputs, it makes standard back-propagation impossible. Existing state-of-the-art methods for network binarization~\citep{courbariaux2015binaryconnect,courbariaux2016binarized,rastegari2016xnor} alternate between a binarized forward pass and a floating point backward pass to circumvent this problem. In their case, the gradients for the $\sign$ activation are approximated during the backward pass, thus introducing inaccuracies in training. Furthermore, batch normalization~\citep{ioffe2015batch} is necessary in binary networks to avoid exploding feature map values due to the large scale of the weights. However, during inference, using batch normalization introduces intermediary floating point representations. This means, despite binarizing weights and activations, these networks can not be used on chips that do not support floating-point computations.

In our method, the scaled hyperbolic tangent function~$\tanh$ is used to bound the values in the range $[-1, 1]$. The network starts with floating point values for weights and activations, and progressively evolves into a binary network as the scaling factor is increased. Firstly, this means that we do not have to toggle between the binary and floating point weight representations. Secondly, we have a continuously differentiable function that allows backpropagation passes. As another important contribution, we reduce the standard batch normalization operation during the inference stage to a simple comparison. This modification is not only very efficient and can be accomplished using fixed-point operations, it is also an order of magnitude faster than the floating-point counterpart. More clearly, while existing binarization methods perform, at each layer, the steps of binary convolutions, floating-point batch normalization, and sign activation, we only need to perform binary convolutions followed by our comparison-based batch normalization, which serves as the sign activation at the same time.

We validate the performance of our self-binarizing networks by comparing them to those of existing binarization methods. We choose the standard bechmarks of CIFAR-10, CIFAR-100~\citep{krizhevsky2009learning} as well as ImageNet~\citep{russakovsky2015imagenet} and popular network architectures such as VGG and AlexNet. We demonstrate higher accuracies despite using less memory and fewer computations. To the best of our knowledge, our proposed networks are the only ones that are free of any floating point computations and can therefore be deployed on low-precision integrated chips or micro-controllers.
 
In what follows, in Sec.~\ref{sec:two} we describe previous work related to reducing the network complexity and where we are placed among them. In Sec.~\ref{sec:three} we explain how the scaled~$\tanh$ function can be used for progressive binarization. In Sec.~\ref{sec:four} we explain our binarization method for weights and activations and explain how to simplify batch normalization at inference time. In Sec.~\ref{sec:five} we compare our technique to existing state-of-the-art binary networks on standard benchmarks and demonstrate the performance gains from our proposed batch normalization simplification. Sec.~\ref{sec:six}~concludes the paper.

\section{Related Work}
\label{sec:two}
Making deep networks memory and computation efficient has been approached in various ways.
In this section we cover some of the relevant literature and explain how our work is positioned with respect to the state-of-the-art. The interested reader may refer to~\citet{cheng_survey_2018}, \citet{ota_survey_2017}, and~\citet{wang_survey_2018} for a wider coverage.

Since most of the computation in deep networks is due to convolutions, it is logical to focus on reducing the computational burden due to them. \citet{howard2017mobilenets,zhang2017shufflenet,freeman2018effnet} employ variations of convolutional layers by taking advantage of the separability of the kernels either by convolving with a group of input channels or by splitting the kernels across the dimensions. In addition to depth-wise convolutions, MobileNetV2~\citep{sandler2018inverted} uses inverted residual structures to learn how to combine the inputs in a residual block. In general, these methods try to design a model that computes convolutions in an efficient manner. This is different from this work because it focuses on redesigning the structure of the network, while ours tries to reduce the memory requirements by using lower precision parameters. However, our method can be applied to these networks as well.

Reducing the number of weights likewise reduces the computational and memory burden. Due to the redundancy in deep neural networks~\citep{cheng2015exploration}, there are some weights that contribute more to the output than others. The process of removing the less contributing weights is called pruning. In~\citet{lecun1990optimal}, the contribution of weights is measured by the effect on the training error when this parameter is zeroed. In Deep Compression~\citep{han2015deep}, the weights with lowest absolute value are pruned and the remaining are quantized and compressed using Huffman coding. In other work, Fisher Information~\citep{tu2016ranking} is used to measure the amount of information the output carries about each parameter, which allows pruning. 
While these approaches often operate on already trained networks and fine-tune them after compression, our method trains a network to a binary state from scratch without removing any parameters or feature maps. This allows us to retain the original structure of the network, while still leaving the potential for further compression after or during binarization using pruning techniques.

Another way to reduce memory consumption and potentially improve computational efficiency is the quantization of weights. Quantized neural networks~\citep{hubara2016quantized,zhou2016dorefa,zhou2017incremental} occupy less memory while retaining similar performance as their full precision counterparts. DoReFaNet~\citep{zhou2016dorefa} proposes to train low bitwidth networks with low bitwidth gradients using stochastic quantization. Similarly,~\citet{zhou2017incremental} devise a weight partitioning technique to incrementally quantize the network at each step. The degree of quantization can vary between techniques. ~\citet{deng2018gxnor},~\citet{zhu2016trained}, and~\citet{fengfu2016ternary} quantize weights to three levels, \emph{i.e.}, two bits only. These quantizations, while severe, still allow for accurate inference. However, to improve computational efficency, specialized hardware is needed to take advantage of the underlying ternary operations. 

An extreme form of such quantization is binarization, which requires only one bit to represent. Expectation BackPropagation (EBP) paved the way for training neural networks for precision limited-hardware~\citep{soudry2014expectation}. BinaryConnect~\citep{courbariaux2015binaryconnect} extended the idea to train neural networks with binary weights. The authors later propose BinaryNet~\citep{courbariaux2016binarized} to binarize activations as well. Similarly, XNORNet~\citep{rastegari2016xnor} extends BinaryNet by adding a scaling factor to the parameters of every layer while keeping the weights and activations binary. ABCNet~\citep{lin2017towards} approximates full precision parameters as a linear combination of binary weights and uses multiple binary activations to compensate for the information loss arising from quantization. \citet{hou2016loss} use Hessian approximations to minimize loss with respect to the binary weights during training.

The focus of our work is the binarization of weights and activations of a network. In previous binarization methods, the binarization process is non-differentiable leading to approximations during the training that can affect the final accuracy. In contrast, we use a differentiable function to progressively self-binarize the network and improve its accuracy. Additionally, we differ from these techniques as we introduce a comparison-based binary batch normalization that eliminates all floating point operations at inference time.

\section{Self-binarization with Tanh}
\label{sec:three}

In typical binary network training, during the forward pass, the floating-point weights and activations are quantized to binary values $\{-1,1\}$, through a piece-wise constant function, most commonly the $\sign$ function:
%
\begin{equation}
	\sign(x) = \begin{cases}
		-1 & \textrm{if } x \leq 0 \\
		1 & \textrm{if } x > 0.
	\end{cases}
\end{equation}
This non-linear activation leads to strong artifacts during the forward pass, and does not generate gradients for backpropagation. The derivatives of the binarized weights are therefore approximately computed using a Straight Through Estimator (STE)~\citep{bengio2013estimating}. STE creates non-zero derivative approximations for functions that either have a zero derivative everywhere or are non-differentiable. Typically, the derivative of $\sign$ is estimated by using the following STE:
\begin{equation}
	\frac{\partial \sign(x)}{\partial x} = \begin{cases}
		1 & \textrm{if } |x| \leq 1 \\
		0 & \textrm{otherwise}.
	\end{cases}
\end{equation}
In the backward propagation step, the gradients are computed on the binarized weights using the STE and the corresponding floating-point representations are updated. Since both forward and backward functions are different, the training is ill-defined. The lack of an accurate derivative for the weights and activations creates a mismatch between the quantized and floating-point values and influences learning accuracy. We term this as \emph{hard binarization}.

This kind of problems has been studied previously, and continuation methods have been proposed to simplify its solution~\citep{allgower2003introduction}. To do so, the original complex and non-smooth optimization problem is transformed by smoothing the original function and then gradually decreasing the smoothness during training, building a sequence of sub-optimization problems converging to the original one. For example, \citet{cao2017hashnet} apply these methods on the last layer of a neural network to predict hashes from images.

\begin{figure}[ht]
	\centering
	\begin{minipage}[b]{0.24\linewidth}
		\centering
		\centerline{\includegraphics[width=\linewidth]{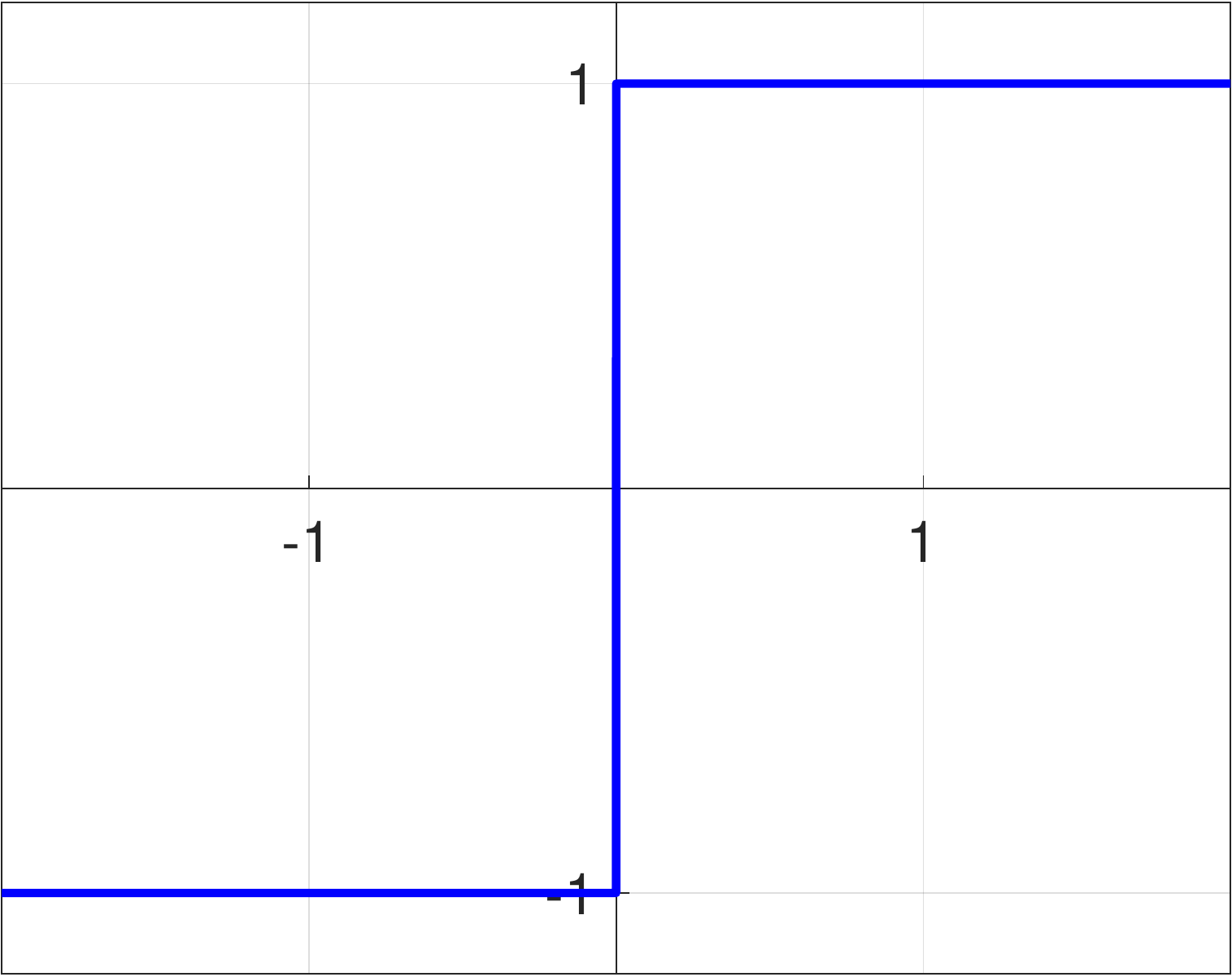}}
		\centerline{(a) $\sign(x)$}
	\end{minipage}
	\hfill
	\begin{minipage}[b]{0.24\linewidth}
		\centering
		\centerline{\includegraphics[width=\linewidth]{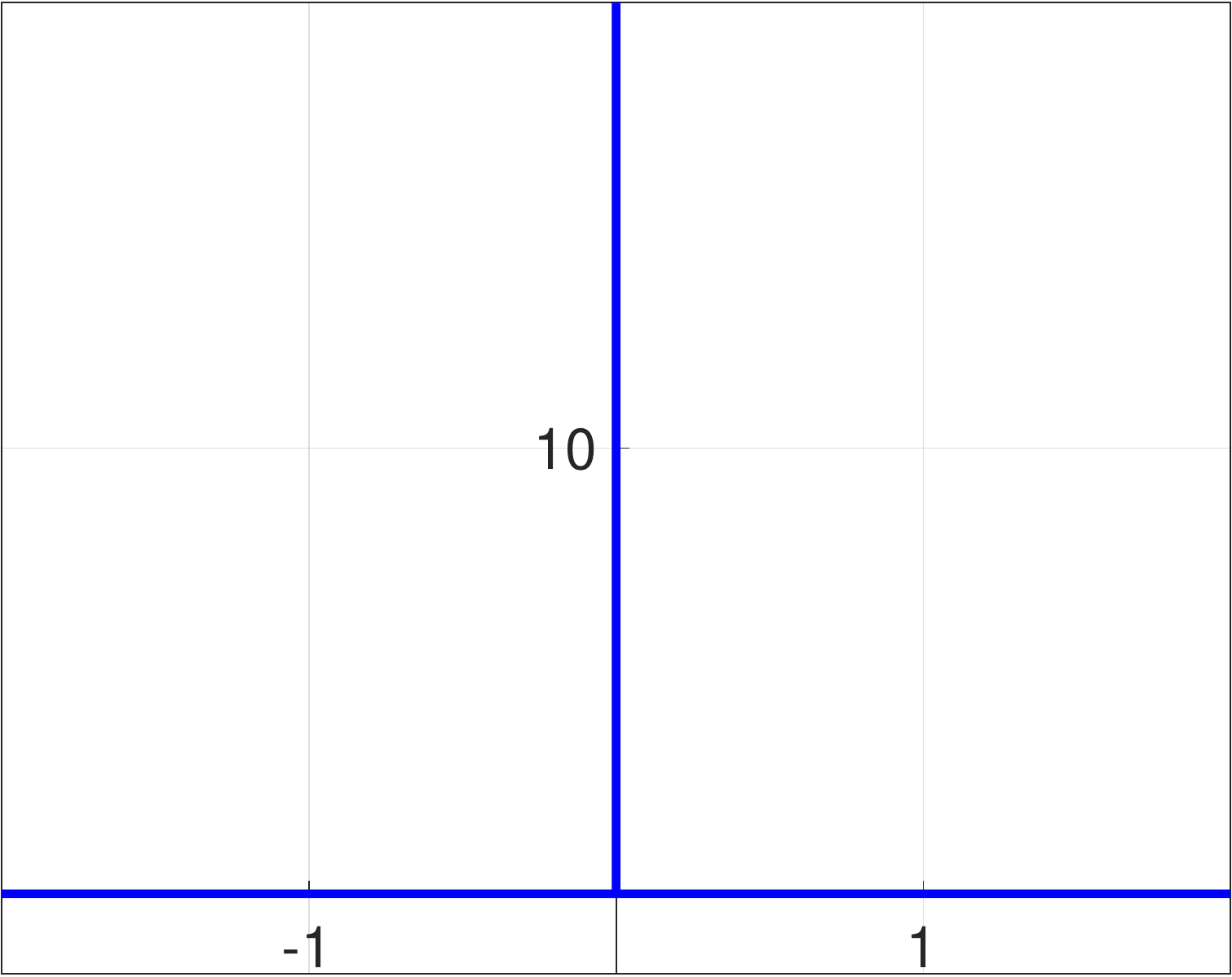}}
		\centerline{(b) $\delta(x)$}
	\end{minipage}
	\hfill
	\begin{minipage}[b]{0.24\linewidth}
		\centering
		\centerline{\includegraphics[width=\linewidth]{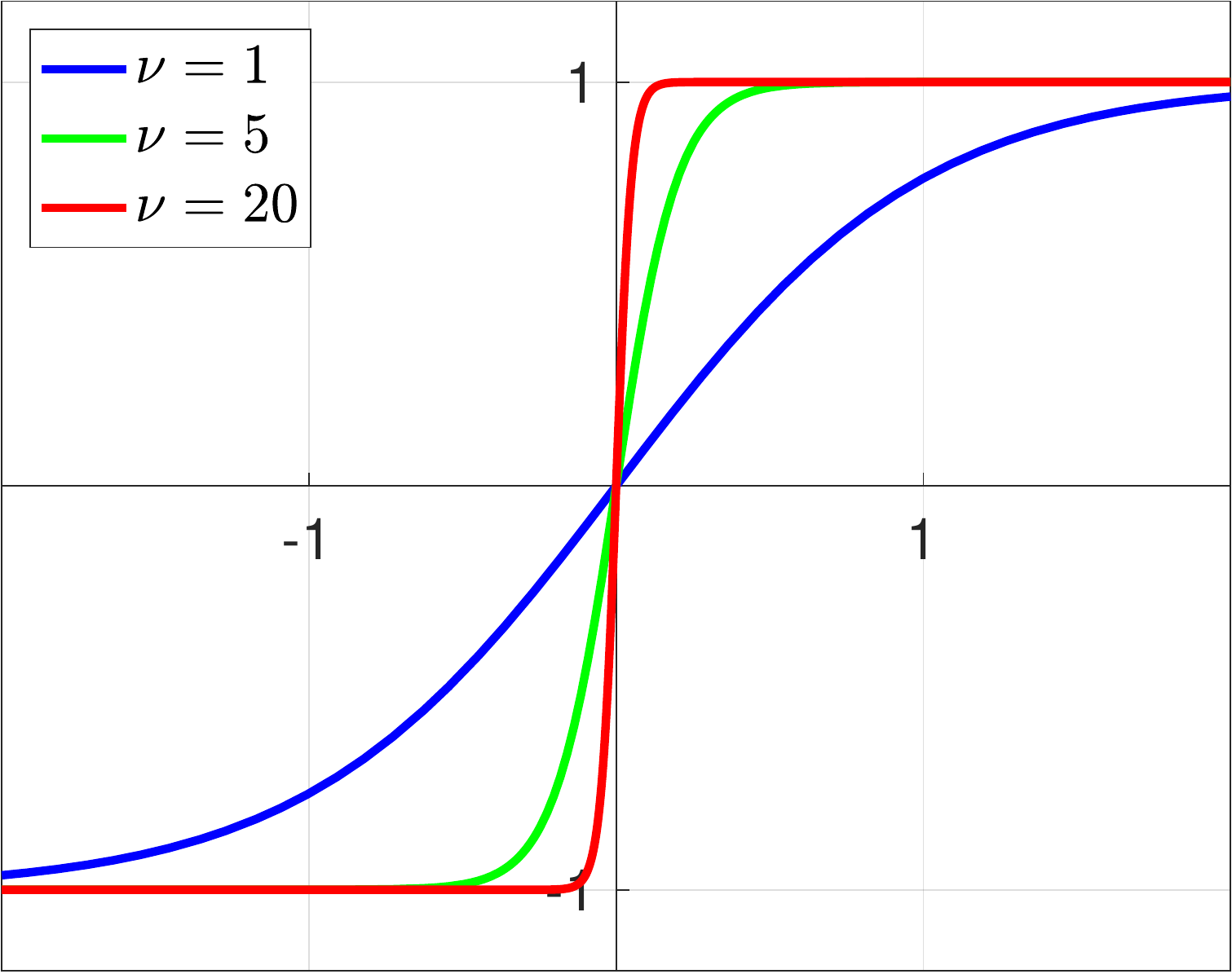}}
		\centerline{(c) $\tanh(\thscale x)$}
	\end{minipage}
	\hfill
	\begin{minipage}[b]{0.24\linewidth}
		\centering
		\centerline{\includegraphics[width=\linewidth]{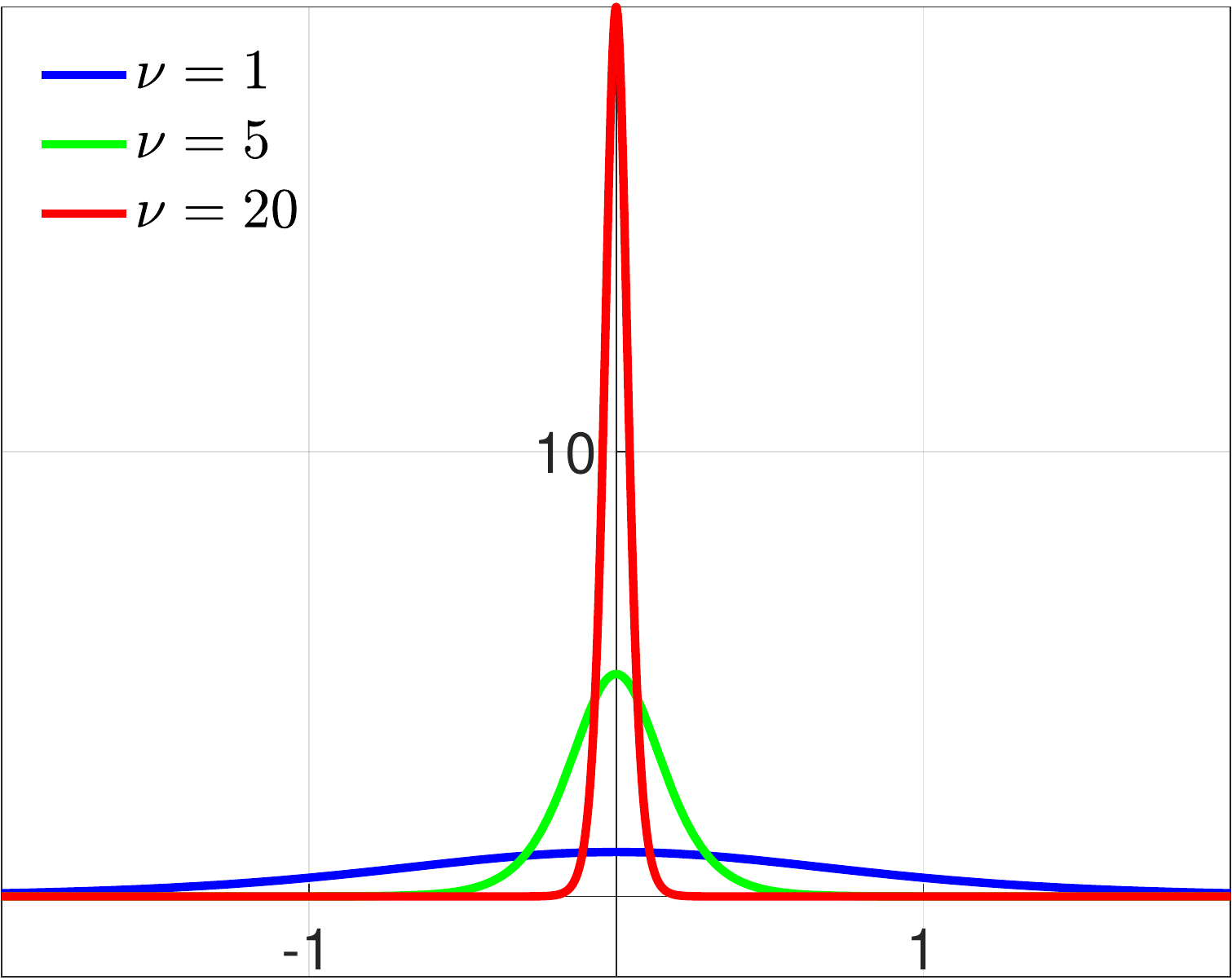}}
		\centerline{(d) $\thscale \text{sech}^2(\thscale x)$}
	\end{minipage}
	\caption{\label{fig:tanh-sign} \emph{The discrete $\sign$ (a) and its derivative (b) are shown. (c, d)~show that as $\thscale$~takes values of~1, 5, and~20, the scaled~$\tanh$ converges toward the original $\sign$ function and its derivative.}}
\end{figure}

Following this philosophy, we introduce a much simpler and efficient training method that allows the network to self-binarize. We pass all our weights and activations through the hyperbolic tangent function $\tanh$ whose slope can be varied using a scale parameter $\thscale > 0$. As seen in Fig.~\ref{fig:tanh-sign}(c), when $\thscale$ is large enough the $\tanh(\thscale x)$ converges to the $\text{sign}(x)$ function:
\begin{equation}
\lim_{\thscale \rightarrow \infty}{\tanh(\thscale x)} = \sign(x).
\end{equation}
Throughout the training process, the weights and activations use floating-point values. Starting from a value of~$\thscale=1$, as the scale factor is progressively incremented, the weights and activations are forced to attain binary values $\{-1, 1\}$. During the training and while $\thscale$ is still small, the derivative of $\tanh$ exists for every value of $x$ and is expressed as
\begin{equation}
\label{eq:tanh-deriv}
\frac{\partial \tanh(\thscale x)}{\partial x} = \thscale \sech^2(\thscale x) = \thscale (1 - \tanh^2(\thscale x)),
\end{equation}
where $\sech$~is the hyperbolic secant.
Using the scaled~$\tanh$, we can build a continuously differentiable network which progressively approaches a binary state, leading to a more principled approach to obtain a binarized network. We term this approach as \emph{soft binarization}.

\section{Method}
\label{sec:four}
%
In this section, we formally describe our self-binarizing approach. We first explain how weights and activations are binarized. Then, we propose a more efficient, comparison-based batch-normalization method that is more suitable when working in binary settings.

\subsection{Weight self-binarization}

As stated above, we cannot use binary weights at training time as it would make gradient computation infeasible. Instead, we use a set of constrained floating-point weights~$\W_\ell$ at each layer~$\ell$. Unlike traditional networks, these weights are not learnable parameters of our model, but depend on learnable parameters. For each layer~$\ell$ of the network, we define a set of learnable, unconstrained parameters~$\P_\ell$ and use the scaled~$\tanh$ to compute the weights as\footnote{For simplicity of notation, we assume that functions act element-wise over vectors and matrices.}
\begin{equation}
    \label{eq:constrained_weights}
    \W_\ell=\tanh(\thscale_e\P_\ell),
\end{equation}
where $\thscale_e$~is the scale factor at epoch~$e$, taken from a sequence~$1 = \thscale_0 < \thscale_1 < \ldots < \thscale_M \rightarrow \infty$ of increasingly larger values. During training, parameters~$\P_\ell$ are updated to minimize a loss function using a standard gradient-based optimization procedure, such as stochastic gradient descent. The scaled~$\tanh$ transforms the parameters~$\P_\ell$ to obtain weights~$\W_\ell$ that are bounded in the range~$[-1, 1]$ and become closer to binary values as the training proceeds.

At the end of the training, weights~$\W_\ell$ are very close to exact binary values. At this point, we obtain the final binary weights~$\B_\ell$ by taking the sign of the parameters,
\begin{equation}
    \label{eq:binary_weights}
    \B_\ell = \sign(\P_\ell) \approx \W_\ell, \quad \forall r.
\end{equation}
At inference time, we drop all learnable parameters~$\P_\ell$ and constrained weights~$\W_\ell$, and use only binary weights~$\B_\ell$.

\subsection{Activation self-binarization}

We follow the idea as for weights to address the binarization of activations as well. During training, we use the scaled~$\tanh$ as the activation function of the network. For a given layer~$\ell$, the activation function transforms the output~$\O_\ell$ of the layer to lead to the activation
\begin{equation}
    \label{eq:activation_tanh}
    \A_\ell = \tanh(\thscale_e\O_\ell).
\end{equation}
The activations are constrained to lie within the range~$[-1, 1]$, and eventually become binary at the end of the training procedure. At inference time we make the network completely binary by substituting the scaled~$\tanh$ by the $\sign$~operator as the binary activation function.

\subsection{Binary Batch Normalization (BinaryBN)}
Batch Normalization (\BN{}) introduced by~\citet{ioffe2015batch} accelerates the training of a general deep network. 
%
During training, the \BN{}~layers compute the running mean $\mu_r$ and standard deviation $\sigma_r$ of the feature maps that pass through them, as well as two parameters, $\beta$ and~$\gamma$, that define an affine transformation. Later, at inference time, \BN{}~layers normalize and transform the input~$I$ to obtain an output~$O$ as given by
\begin{equation}
    O = \dfrac{I - \mu_r}{\sigma_r} \gamma + \beta.
\end{equation}

For binary networks~\citep{courbariaux2015binaryconnect,courbariaux2016binarized,rastegari2016xnor} in particular, \BN{}~becomes essential in order to avoid exploding activation values. However, using~\BN{} brings in the limitation of relying on floating-point computations. Apart from affecting computation and memory requirements, the floating-point \BN{} effectively eliminates the possibility of using the network on low-precision hardware.
%

A useful observation of \BN{} is that, in a networks with binary activations, the output of the \BN{}~layers is always fed into a $\sign$~function. This means only the sign of the normalized output is kept, while its magnitude is not relevant for the subsequent layers. We can leverage this property to simplify the \BN{} operation. 
The sign of the output~$O$ of a \BN{} layer can be reformulated as:
\begin{equation}
\label{eq:binbn}
\begin{split}
\sign(O) & \equiv O > 0 \equiv \frac{I - \mu_r}{\sigma_r} \gamma + \beta > 0 \\
 & \equiv (I - \mu_r) \gamma > - \sigma_r\beta \\
 & \equiv I \sign(\gamma) > \left(\mu_r - \frac{\sigma_r\beta}{\gamma}\right)\sign(\gamma) \quad \text{since dividing by}\ \gamma\ \text{could flip the inequality} \\
 & \equiv I \sign(\gamma) > T \sign(\gamma) \\
 & \equiv \text{XNOR}(I > T, \gamma > 0),
\end{split}
\end{equation}
with
\begin{equation}
T = \mu_r - \frac{\sigma_r \beta}{\gamma}.
\end{equation}
While $T$~is a floating-point value, in practice we represent it as a fixed-point 8-bit integer. This sacrifices some amount of numerical precision, but we observed no negative effect on the performance of our models. We refer to our simplified batch normalization as Binary Batch Normalization (\BinaryBN{}).

Note that the derivation of Eq.~\eqref{eq:binbn} does not hold when $\gamma=0$. This is handled as a special case that simply evaluates~$\beta > 0$.  It must be emphasized that \BinaryBN{} is not an approximate method; it computes the exact value of the sign of the output of the standard~\BN{}.

During training we use the conventional \BN{}~layers. At inference time, we replace them with the \BinaryBN{} without any loss in prediction accuracy. Our trained models can thus bypass the floating-point batch normalization with an efficient alternative that requires mere comparison operations.



\section{Experiments}
\label{sec:five}
In this section, we first compare our self-binarization method against other known techniques. Later on, we discuss and demonstrate the efficiency gained using our proposed BinaryBN instead of the typical BN~layer.

\subsection{Classification Accuracy}

We compare our self-binarizing networks with other state-of-the-art methods that use binary networks. ~\citet{courbariaux2015binaryconnect} present \emph{BinaryConnect} (\BC), a method that only binarizes the weights. \emph{Binary Neural Networks}~(\BNN)~\citep{courbariaux2016binarized} improves~\BC \,by also binarizing the activations. Similarly, \citet{rastegari2016xnor} present two variants of their method: \emph{Binary Weight Networks}~(\BWN), which only binarizes the weights, and \emph{XNORnet} (\XNOR), which binarizes both weights and activations.

For a fair comparison and to prove the generality of our method, we use the original implementations of \BC, \BNN, \BWN, and \XNOR, and apply our self-binarizing technique to them. Specifically, we substituted the weights from the original implementations by a pair of parameters~$\P$ and constrained weights~$\W$ given by Eq.~\eqref{eq:constrained_weights}. Also, we substituted the activation functions by the scaled~$\tanh$ as described in Eq.~\eqref{eq:activation_tanh}. At inference time we used the binary weights obtained with Eq.~\eqref{eq:binary_weights} and the $\sign$~operator as the activation function. Additionally, we replace the batch normalization layers by our BinaryBN layer for the cases where the activations are binarized.

We evaluate the methods on three common benchmark datasets: CIFAR-10, CIFAR-100~\citep{krizhevsky2009learning} and ILSVRC12 ImageNet~\citep{russakovsky2015imagenet}. For CIFAR-10 and CIFAR-100, we use a VGG-16-like network with data augmentation as proposed in~\citet{lee2015deeply}: 4 pixels are padded on each side, and a 32x32 patch is randomly cropped from the padded image or its horizontal flip. During testing, only a single view of the original 32x32 image is evaluated. The model is trained with a mini-batch size of~256 for 100~epochs.

The ILSVRC12 ImageNet dataset is used to train an AlexNet-like network without drop-out or local response normalization layers. We use the data augmentation strategy from~\cite{wu2016tensorpack}. At inference time, only the center crop is used from the validation set. The model is trained with a mini-batch size of~64 and a total of 50~epochs.

\newcommand{\BINW }{\rotatebox[origin=c]{90}{\tiny \textbf{\makecell{WEIGHT-ONLY\\BINARIZATION}}}}
\newcommand{\BINWA}{\rotatebox[origin=c]{90}{\tiny \textbf{\makecell{WEIGHT \& \\ACTIVATION\\BINARIZATION}}}}
%
%

In all our experiments we increase $\thscale$ from~$1$ to~$1000$ during training in an exponential manner. The final $\thscale=1000$ is large enough to make weights and activations almost binary in practice. We optimize all models using Adam~\citep{adam_kingma_2014} with an exponentially decaying learning rate starting at $10^{-3}$.


\begin{table}[ht]
	\begin{center}
		\caption{\label{tbl:binary-bn-comparison}\emph{Accuracy rates (\%) comparing existing state-of-the-art methods when trained with and without our self-binarizing technique. Top-1 and Top-5 results are reported for CIFAR-100 and ImageNet. Also, Top-1 results are reported for CIFAR-10. $B_w$, $B_a$, and $B_{BN}$ indicate the number of bits required to represent the values of the weights, activations, and batch normalization outputs.
		Note that since we use BinaryBN, which combines the batch normalization and activation steps, we do not show any bits for~$B_a$ when both weights and activations are binarized using our method.}}
		\begin{tabular}{r | l l | c c c | c c c}

		\hline\noalign{\smallskip}
		& \textbf{Method} & & $B_w$ & $B_a$ & $B_{BN}$  & CIFAR-10 & CIFAR-100 & ImageNet \\
		\noalign{\smallskip} \hline\noalign{\smallskip}
		& \textbf{Full-Precision} & & 32 & 32 & 32 & 86.55 & 60.67 / 81.98 & 55.07 / 79.27 \\
		\noalign{\smallskip} \hline\noalign{\smallskip} 
		
		\multirow{4}{*}{\BINW} & \multirow{2}{*}{\textbf{BC}} & orig. & 1 & 32 & 32  & 87.24 & \textbf{55.82} / 73.11 & 40.57 / 66.40 \\
		& & ours & 1 & 32 & 32 & \textbf{88.50} & 55.81 / \textbf{76.50} & \textbf{41.27 / 67.51} \\
		\noalign{\smallskip}
		
		& \multirow{2}{*}{\textbf{BWN}} & orig. & 1 & 32 & 32 & 89.87 & 63.14 / 85.64 & 50.87 / 75.63 \\
		& & ours & 1 & 32 & 32 & \textbf{90.56} & \textbf{63.48 / 86.01} & \textbf{52.89 / 77.34} \\
		\noalign{\smallskip}\hline\noalign{\smallskip}
		
		\multirow{4}{*}{\BINWA} & \multirow{2}{*}{\textbf{BNN}} & orig. & 1 & 1 & 32 & 84.06 & 51.58 / 78.58 & 29.94 / 56.43 \\
		& & ours & 1 & - & 1 & \textbf{84.61} & \textbf{54.06 / 80.45} & \textbf{30.97 / 56.65} \\
		\noalign{\smallskip}
		
		& \multirow{2}{*}{\textbf{XNOR}} & orig. & 1 & 1 & 32 & 86.16 & 51.91 / 79.41 & 37.84 / 64.06 \\
		& & ours & 1 & - & 1 & \textbf{86.91} & \textbf{53.84 / 81.56} & \textbf{39.44 / 65.23} \\
		\noalign{\smallskip}\hline\noalign{\smallskip}
		\end{tabular}
	\end{center}
\end{table}
%
%

Table~\ref{tbl:binary-bn-comparison} shows the results of the experiments. Our self-binarizing approach achieves the highest accuracies for CIFAR-10 and ImageNet. Our method is only slightly outperformed by~\BC{} for CIFAR-100, but still gives better Top-5 accuracy for the same model. For both weights and activation binarization, our method obtains the best results across all datasets and architectures.

What is remarkable is that the improved performance comes despite eliminating floating-point computations and using drastically fewer bits than the previous methods, as can be seen in the columns~$B_w$, $B_a$ and $B_{BN}$ of Table~\ref{tbl:binary-bn-comparison}.
For CIFAR-10 and CIFAR-100, \BWN~outperforms the full precision models likely because the binarization serves as a regularizer~\citep{courbariaux2015binaryconnect}.


\subsection{Efficiency due to BinaryBN}


With all other computations being common between our self-binarizing networks and other networks with binary weights and activations, any difference in computational efficiency has to arise from the use of a different batch normalization scheme. We therefore compare our proposed \BinaryBN{} layer to the conventional \BN{} layer as well as to the Shift-based Batch Normalization (\SBN{}) proposed by~\citet{courbariaux2016binarized}. \SBN{}~proposes to round both~$\gamma$ and~$\sigma_r$ to their nearest power of~2 and replace multiplications and divisions by left and right shifts, respectively. \SBN{}~follows the same formula as~\BN{}, and is used both at training and inference time, so that the network is trained with the rounded parameters.

\begin{table}[ht]
	\begin{center}
		\caption{\label{tbl:batch-norm-bin} \emph{Memory and computational requirements of BN, SBN and BinaryBN~layers when working on a $w\times h$~feature map with $c$~channels.}}
		\begin{tabular}{l | c c c}
		\hline\noalign{\smallskip}
		& \BN{} + $\sign$ & \SBN{} + $\sign$ & \BinaryBN{} \\ 
		\noalign{\smallskip} \hline\noalign{\smallskip} 
		Operations used   & $+, -, \times, \div$ & $+, -, \ll, \gg$ & $>, =$ \\
		Storage Memory  & $128c$ & $80c$ & $9c$ \\
		Output Memory & $32chw + chw$ & $32chw + chw$ & $chw$ \\
		Operations & $4chw + chw$ & $4chw + chw$ & $2chw$ \\
		\noalign{\smallskip}\hline\noalign{\smallskip}
		\end{tabular}
	\end{center}
\end{table}

Table.~\ref{tbl:batch-norm-bin} summarizes the requirements of memory and computational time of these three types of batch normalization layers. We assume the standard case where a binary convolution is followed by a batch normalization layer and then a binary activation function.

For storing \BN{} parameters, we need four 32-bit vectors of length~$c$, amounting to $32\times4c = 128c$\,bits, $c$~being the number of channels in this layer. For \SBN{}, we need two 32-bit vectors and two 8-bit vectors of length~$c$, resulting in $80c$\,bits. For \BinaryBN{}, we need an 8-bit vector of size~$c$ to store the $T$~value of each channel, and a binary vector for the sign of~$\gamma$, totaling $9c$\,bits.


\begin{figure}[ht]
	\centering
	\begin{minipage}[b]{0.39\linewidth}
		\centering
		\centerline{\includegraphics[width=\linewidth]{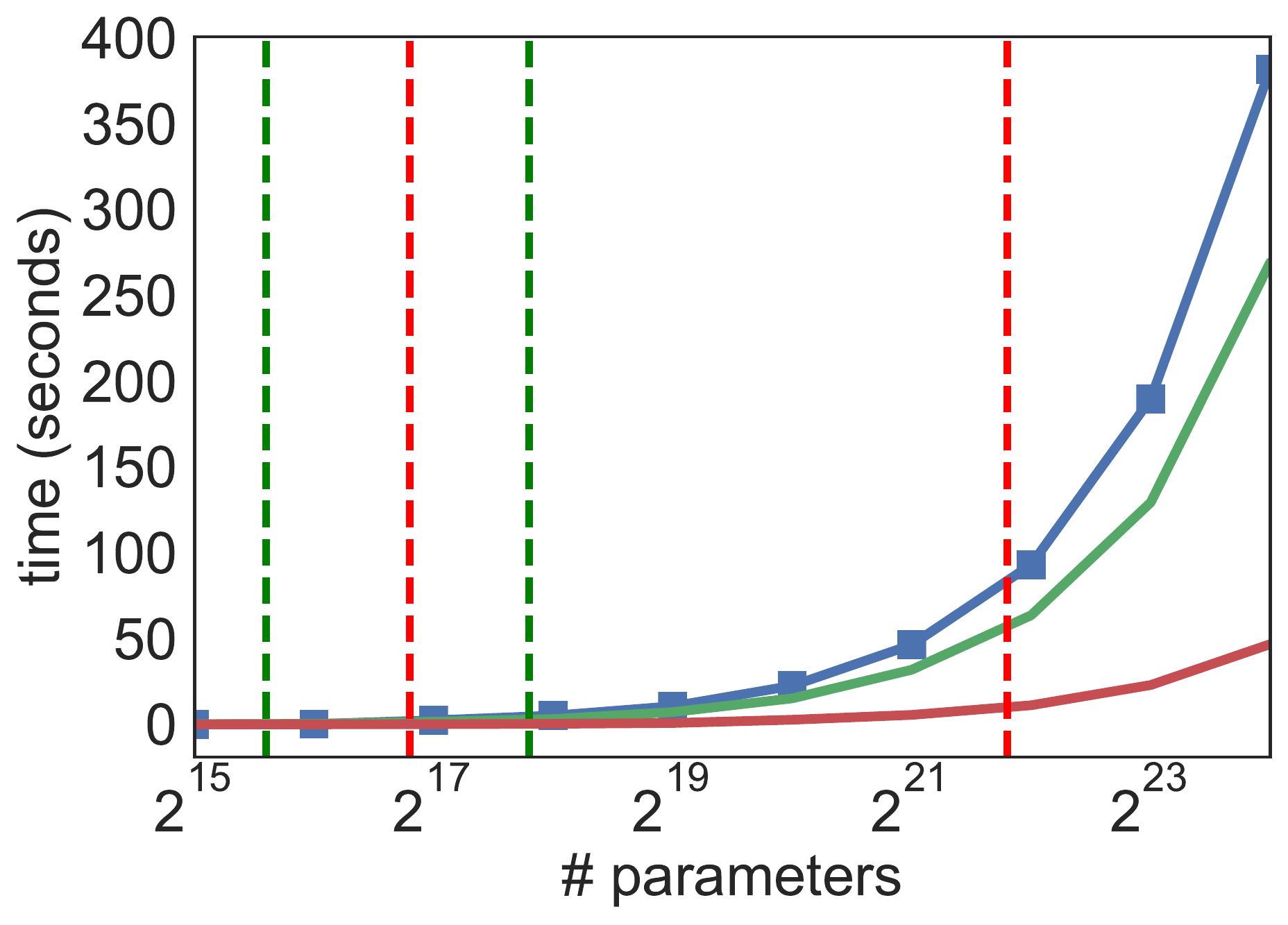}}
		\centerline{(a) Time comparison}
	\end{minipage}
	\hfill
	\begin{minipage}[b]{0.19\linewidth}
		\centering
		\centerline{\includegraphics[width=\linewidth,trim=0 -2.5cm 0 0]{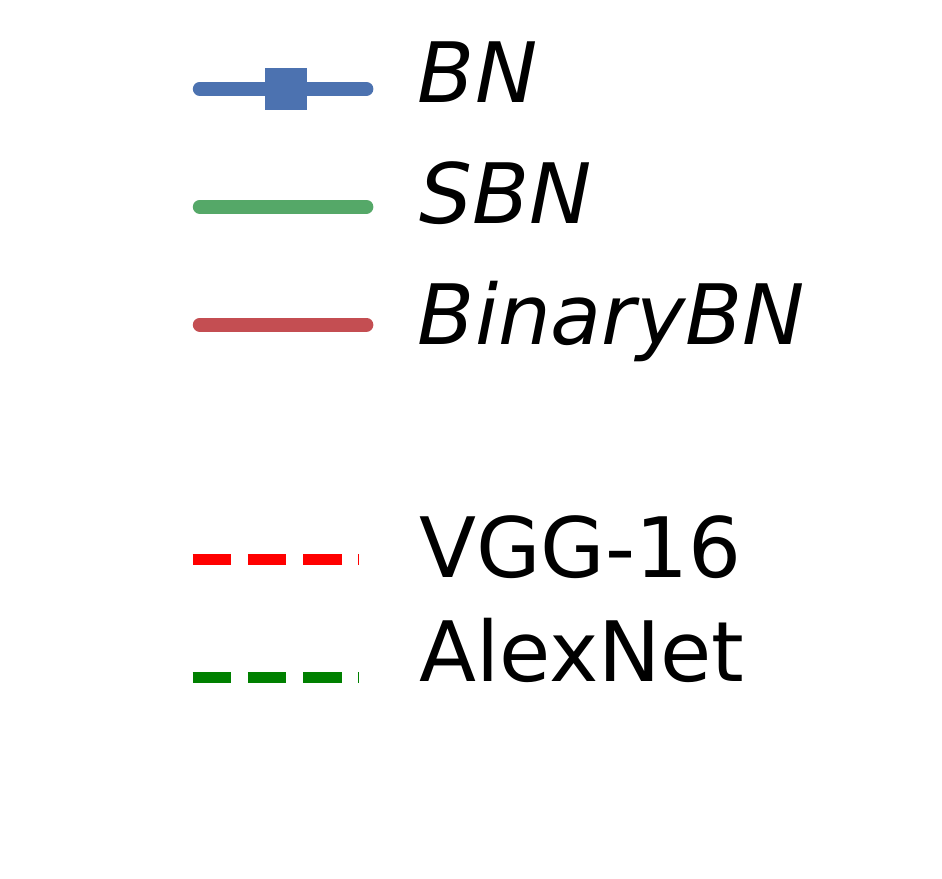}}
	\end{minipage}
	\hfill
	\begin{minipage}[b]{0.39\linewidth}
		\centering
		\centerline{\includegraphics[width=\linewidth]{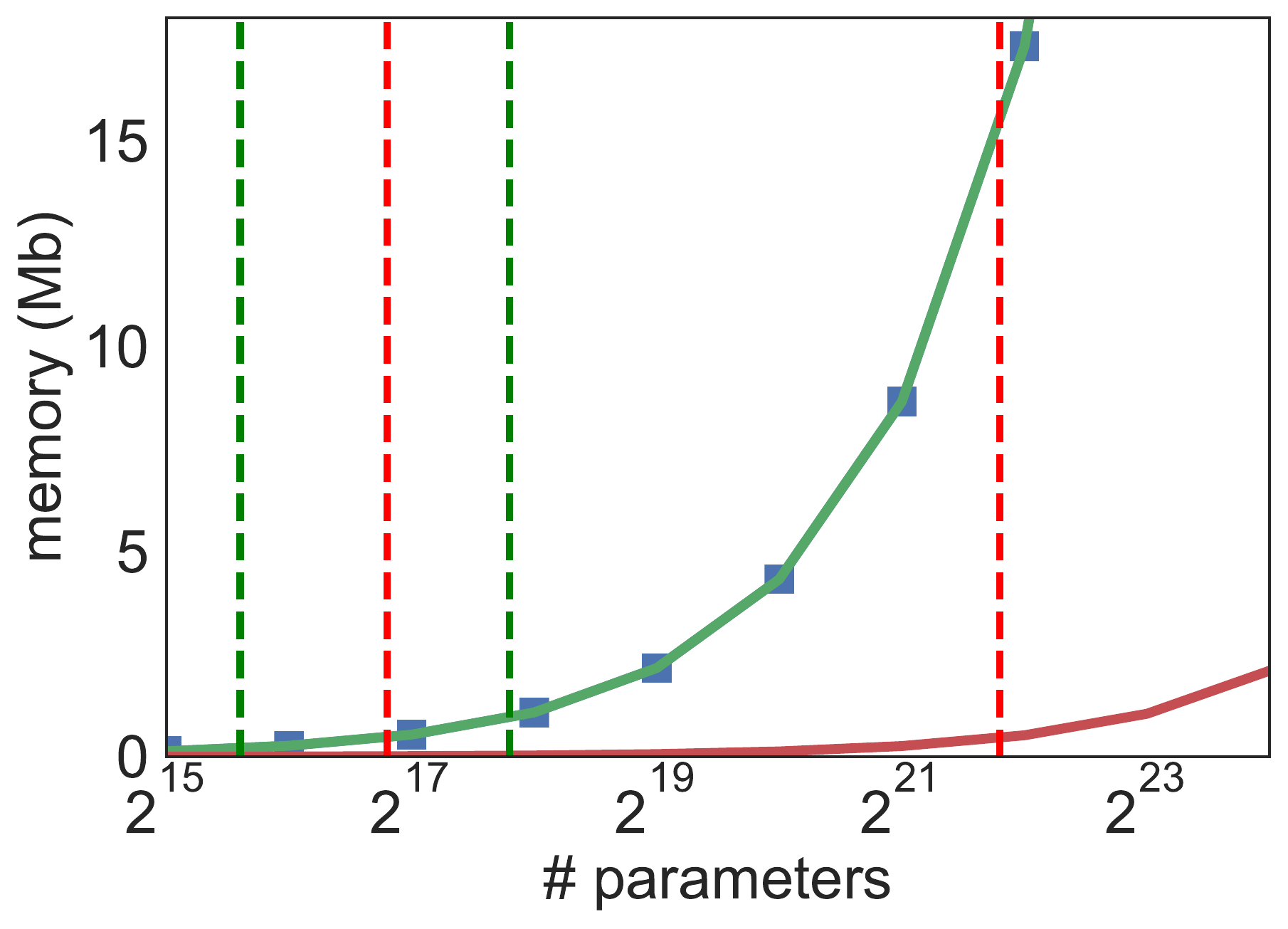}}
		\centerline{(b) Memory comparison}
	\end{minipage}
	\caption{\label{fig:bn-time} \emph{Comparison between time and memory consumption of BN, SBN and BinaryBN for an input feature map of $256 \times 256$. For reference, dotted lines show the minimum and maximum number of parameters that can be found in the BN layers of a VGG-16 network and an AlexNet network.}}
\end{figure}

We experimentally assessed the time and memory requirements of the three batch normalization techniques. We run batches of increasing sizes through \BN{}, \SBN{} and \BinaryBN{} layers with randomly generated values for $\mu_r$, $\sigma_r$, $\beta$, and~$\gamma$, and measure time and memory consumption. Fig.~\ref{fig:bn-time} shows the results. Overall, \BinaryBN{} is nearly one order of magnitude less memory consuming and faster than \BN{} and~\SBN{}.


\subsection{Why soft binarization is better}
\begin{figure}[ht]
	\centering
	\begin{minipage}[b]{\linewidth}
		\centering
		\centerline{\includegraphics[width=\linewidth]{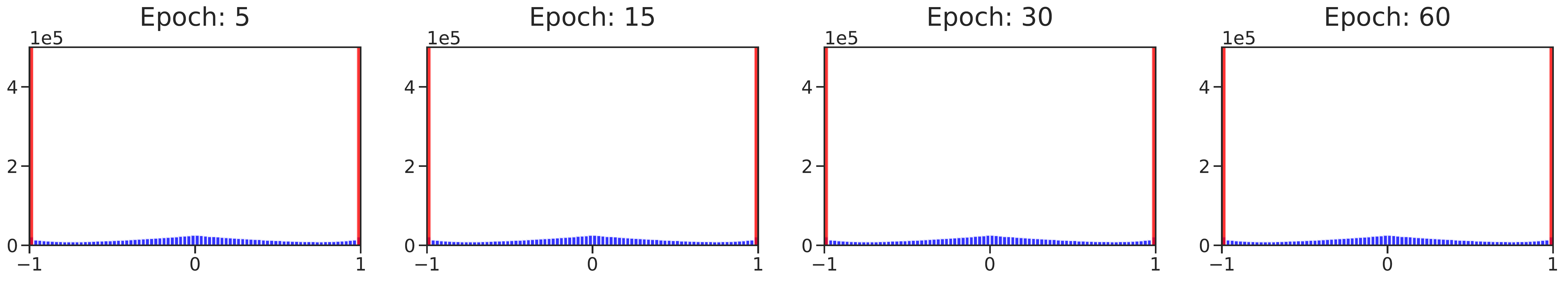}}
		\centerline{(a) Hard binarization with $\text{sign}(x)$}
		\medskip
	\end{minipage}
	\begin{minipage}[b]{\linewidth}
			\centering
			\centerline{\includegraphics[width=\linewidth]{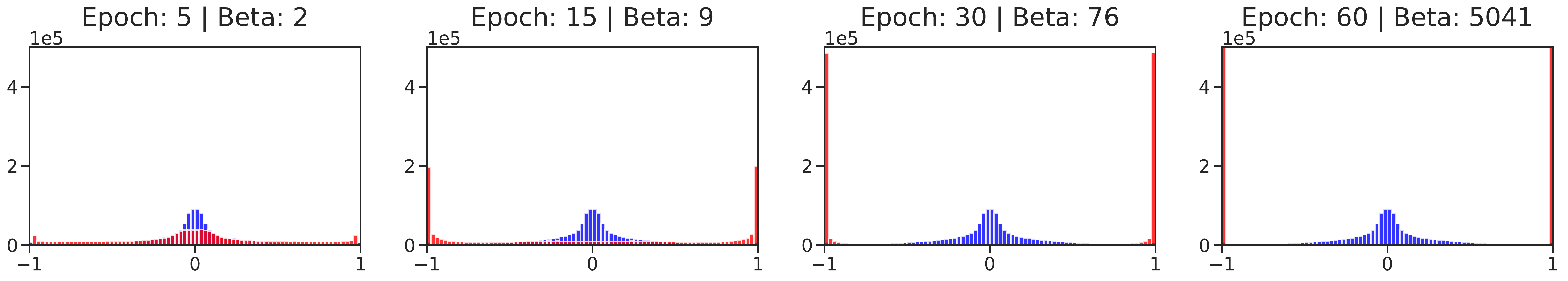}}
			\centerline{(b) Soft binarization with $\tanh(\thscale x)$}
		\end{minipage}
	\caption{\label{fig:snapshots} \emph{Histograms of weights before (blue) and after (red) binarization for conventional approaches in (a) versus our method in (b).}}
\end{figure}

In order to show the superiority of our soft binarization based on scaled $\tanh$ over the hard binarization of \BC, \BNN, \BWN, and \XNOR, we take snapshots of the weights at different epochs of training a VGG-16 network on CIFAR10. Fig.~\ref{fig:snapshots} shows the histograms of snapshots taken at the last convolutional layer of the network. It can be seen that, under the soft binarization scheme, the floating-point weights are initially centered at $0$, allowing them to continue evolving as the network progresses. As the training proceeds, the distribution of softly-binarized weights is gradually shifted towards binary values. However, under the hard binarization scheme, the distribution of floating-point weights is not similarly centered at zero, which means a lot of the weights have immediately been driven far from $0$ and require very large gradients to be changed. With weights unable to change as much, a lot of the neurons in the network become quickly unusable thereby limiting the final performance of the architecture. This explains why our binary networks perform better than those relying on hard binarization.

\section{Conclusion}
\label{sec:six}
We present a novel method to binarize a deep network that is principled, simple, and results in binarization of weights and activations. Instead of relying on the $\sign$ function, we use the $\tanh$ function with a controllable slope. This simplifies the training process without breaking the flow of derivatives in the back-propagation phase as compared to that of existing methods that have to toggle between floating-point and binary representations. In addition to this, we replace the conventional batch normalization, which forces existing binarization methods to use floating point computations, by a simpler comparison operation that is directly adapted to networks with binary activations. Our simplified batch normalization is not only computationally trivial, it is also extremely memory-efficient. Despite using lesser memory and computation, our trained binary networks outperform those of existing binarization schemes on the standard benchmarks.


\bibliography{egbib,egbib2}
\bibliographystyle{iclr2019_conference}


\end{document}